\documentclass[letterpaper,10pt,conference]{IEEEtran}

\IEEEoverridecommandlockouts

\ifdefined\pdfobjcompresslevel
\fi

\usepackage{amsmath,amssymb}
\usepackage{algorithm}
\usepackage{algpseudocode}
\usepackage{array}
\usepackage{booktabs}
\usepackage{cite}
\usepackage{makecell}
\usepackage{microtype}
\usepackage{multirow}
\usepackage{tabularx}
\usepackage{graphicx}
\usepackage{xcolor}

\usepackage{tikz}
\usetikzlibrary{arrows.meta, positioning, fit}

\usepackage[font=small,labelfont=bf]{caption}
\usepackage{hyperref}
\usepackage{url}

\usepackage[letterpaper,left=0.77in,right=0.77in,top=0.76in,
            bottom=0.84in,columnsep=0.2in]{geometry}
\renewcommand{\IEEEtitletopspaceextra}{12pt}
\microtypesetup{protrusion=false}
\hypersetup{hidelinks}
\newcolumntype{Y}{>{\raggedright\arraybackslash}X}
\newcolumntype{L}[1]{>{\raggedright\arraybackslash}m{#1}}
\newcolumntype{C}[1]{>{\centering\arraybackslash}m{#1}}

\newcommand{\C}{\mathcal}

\definecolor{mygray}{gray}{0.6}

\usepackage{url}

\title{%
    \LARGE
    \textbf{Search, Ground, Plan:}
    Functional Sufficiency for Task and Motion Planning under Incomplete Scene Knowledge
}

\author{
\IEEEauthorblockN{
Narendhiran Vijayakumar\textsuperscript{*1},
Nav Singhal\textsuperscript{1},
Girish Varma\textsuperscript{2},
Antony Thomas\textsuperscript{1}
}
\IEEEauthorblockA{
\textsuperscript{1}Robotics Research Center, IIIT Hyderabad
\hspace{0.025em}
\textsuperscript{2}Center for Security, Theory and Algorithmic Research, IIIT Hyderabad
}
}
\begin{document}

\setcounter{page}{1}

%


\IEEEaftertitletext{%
    \vspace{-0.6cm}
    \begin{center}

        \includegraphics[
            width=0.97\textwidth,
            trim={0.2cm 9.45cm 3cm 0.05cm},
            clip
        ]{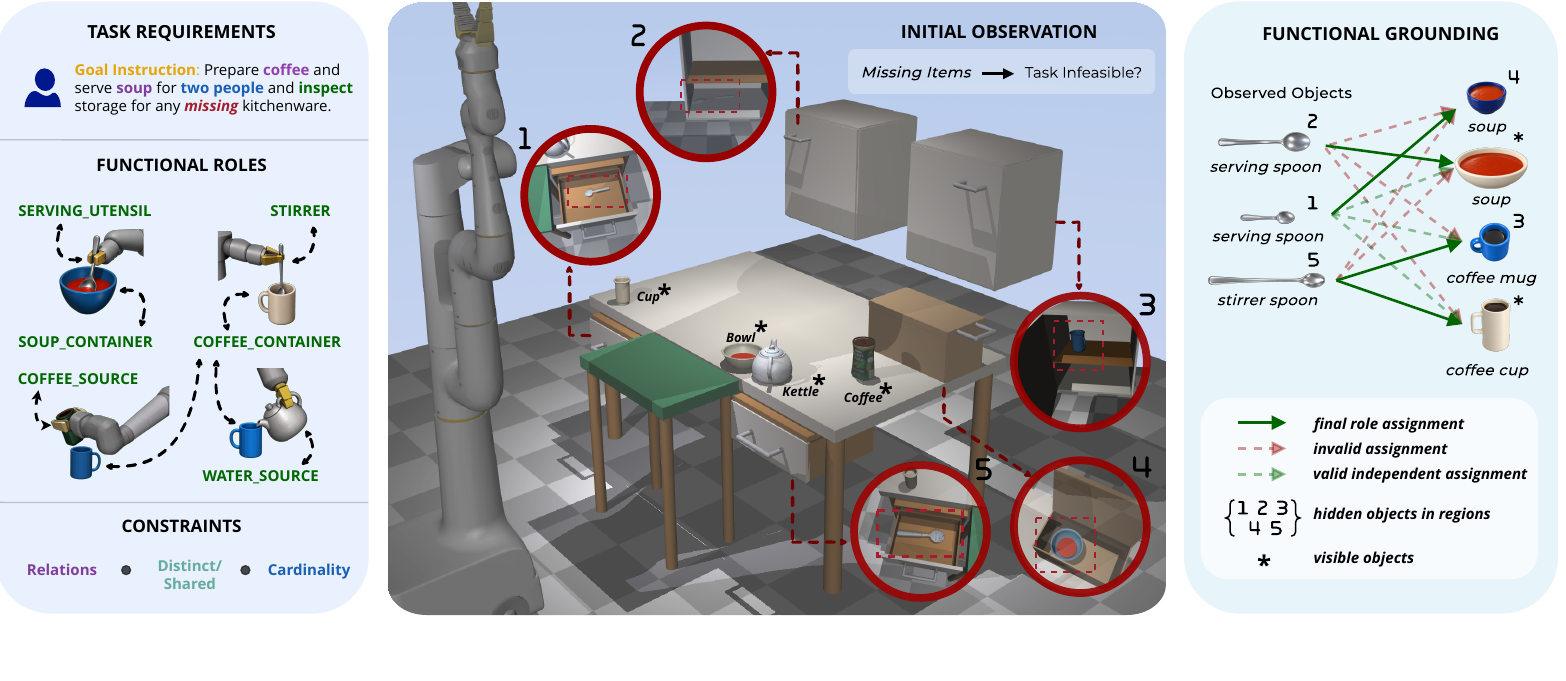}

        \refstepcounter{figure}
        \label{fig:teaser}

        \vspace{-0.35cm}

        \begin{minipage}{0.99\textwidth}
          \vspace{-0.35cm}
            \centering
            \footnotesize
            Fig.~\thefigure.
            A task and motion planning problem with incomplete object details,
            inter-object relations, and scene observations.
            \textsc{GRAB-TAMP} incorporates task-level functional requirements
            (left), searches for hidden task-relevant objects (center), and verifies
            joint role assignments (right) before planning.
        \vspace{0.1cm}
        \end{minipage}

    \end{center}
}
\maketitle

\begin{abstract}
~Foundation models (FMs) have expanded task and motion planning (TAMP) to manipulation problems specified through language and visual observations. However, incomplete scene knowledge leaves a critical gap between understanding what the task requires and knowing whether the physical scene can actually realize it. We introduce \textbf{{GRAB-TAMP}}, an FM-based TAMP framework that \textit{searches} for scene entities required for task completion, \textit{grounds} functional roles to valid physical objects, and \textit{plans} only after a complete joint assignment establishes functional sufficiency. We represent the task through functional roles, relations, and assignment constraints, and incrementally inspect the scene while requirements remain unresolved, verifying candidate objects through semantic, geometric, and relational checks. We evaluate GRAB-TAMP across 32 scene variants spanning Kitchen, Living Room, and Workshop domains. Across 200 feasible trials, our approach achieves \(54.0\%\) end-to-end success with \(67.3\%\) plan goal coverage. Compared with three FM-based TAMP frameworks under the same execution setting, GRAB-TAMP improves end-to-end success by \(25.7\) percentage points over the mean baseline. Implementation and evaluation code: \url{https://anonymous.4open.science/r/GRAB-TAMP}.
\end{abstract}


\section{Introduction}
\vspace{-0.15cm}
Many everyday manipulation tasks appear trivial largely because humans implicitly reason about far more than what is explicitly stated in the instruction. An instruction specifies the intended outcome, but rarely describes every object, relation, or intermediate observation required to achieve it. Consider the scene in Fig.~\ref{fig:teaser}, where the robot is instructed to \emph{prepare coffee and serve soup for two people}. The language instruction inherently specifies the functional requirements for task completion, but leaves unresolved which physical scene objects satisfy them. This becomes challenging when the initial observation does not provide enough evidence to make that realization. For example, a suitable object may be occluded, contained within an articulated region, or simply outside the present view. Its absence from the current observation therefore does not imply that the task itself is infeasible. 

At the same time, a new observation alone may be insufficient. A newly observed object may be irrelevant to the task, semantically suitable but geometrically incompatible, or satisfy both requirements. For example, a spoon may be recognized as an appropriate utensil but too large for the intended container, rendering it geometrically incompatible (see functional grounding block in Fig.~\ref{fig:teaser}). Thus, a valid functional assignment must satisfy both the semantic and geometric requirements of the role it fulfills. 


Fig.~\ref{fig:teaser} further illustrates why certain functional assignments must be considered jointly. A narrow spoon may fit both a small and a wide bowl, while a larger spoon may fit only the wide bowl. Assigning the narrow spoon to the wide bowl may thus appear locally valid but leave no feasible utensil for the small bowl. This shows why object--role assignments must be considered jointly rather than independently. We refer to this problem as \textbf{\textit{functional grounding}}, assigning physical scene entities to task-level roles subject to the properties, relations, cardinalities, and interaction requirements imposed by the task. Whether an object constitutes a valid choice may depend on what has already been assigned in the task. Conversely, the absence of a valid assignment in the current observation does not necessarily imply that the task itself is infeasible. 

Given a functional grounding, a Task and Motion Planning (TAMP) framework~\cite{garrett2021integrated} can reason about the discrete actions and the continuous motions required to execute them. Classical TAMP methods like PDDLStream \cite{garrett2020pddlstream}, combines discrete decisions with continuous reasoning over grasps, placements, configurations, and motions. Recent embodied language ~\cite{huang2022zeroshot} and multimodal models \cite{huang2023inner} have substantially improved the ability of robots to interpret abstract natural-language instructions, incorporate visual context, and reason over long-horizon tasks. Yet, under incomplete scene knowledge, a fundamental question remains unresolved: \emph{whether the available physical entities are functionally sufficient to fulfill the task requirements before downstream planning and execution.}

This leaves a gap between interpreting task requirements and determining which scene entities satisfy those requirements for achieving \textbf{\textit{task feasibility}}. In our framework, we
represent the former as a fixed functional specification derived from the task,
while the latter is resolved from the scene evidence accumulated thus far. When the currently observed objects are insufficient, additional scene information is acquired and newly revealed candidates are checked against the task requirements. Once a complete, valid object-role assignment is established, a TAMP planner can then verify task feasibility without repeatedly generating and evaluating partial plans. 

\begin{figure*}[t]
    \centering
    \includegraphics[
        width=0.93\linewidth,
        trim={2.1cm 7.6cm 0.75cm 0.3cm},
        clip
    ]{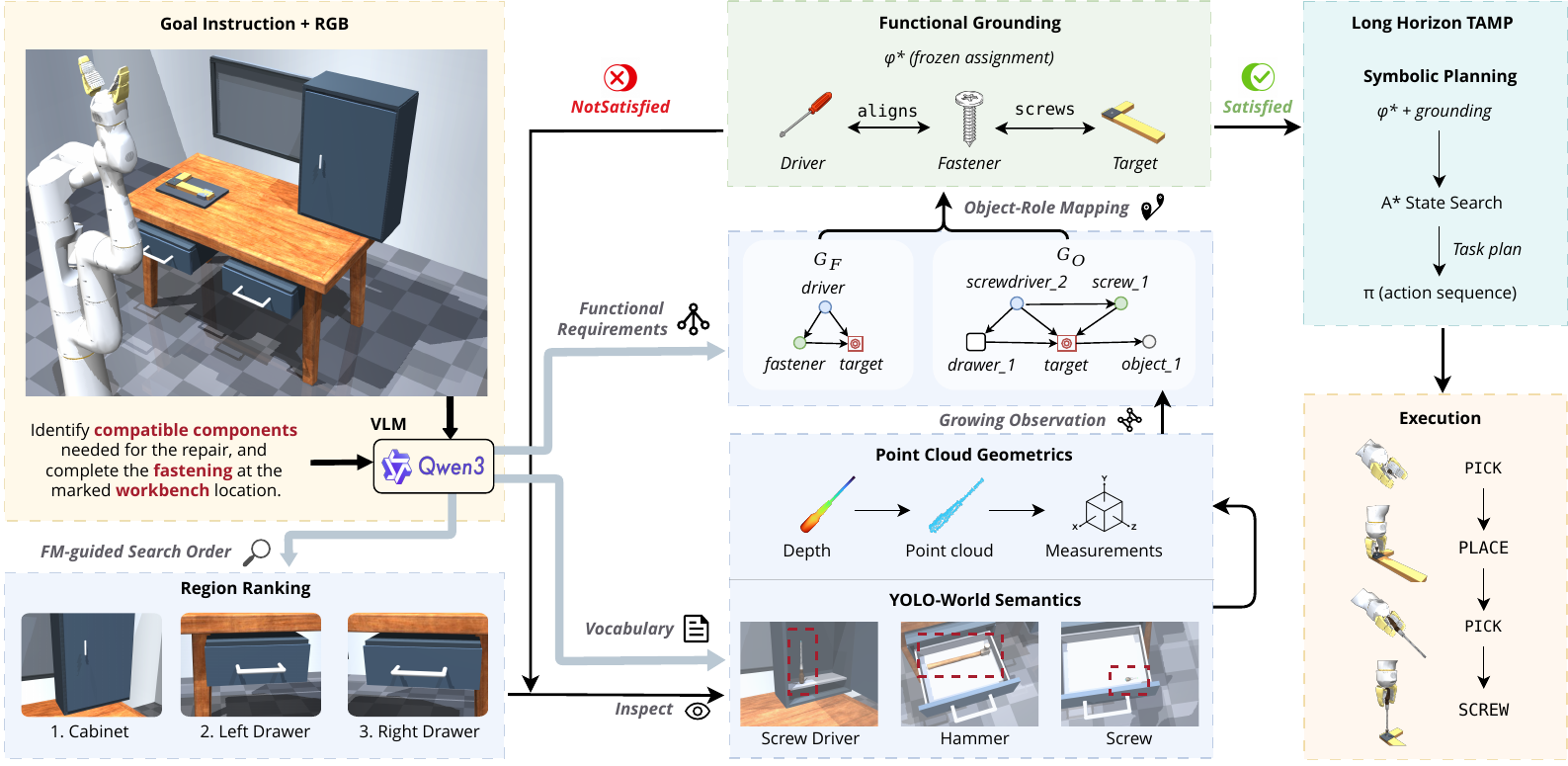}
    \vspace{-0.23cm}
    \caption{\textsc{GRAB-TAMP} uses an FM to convert the goal and initial RGB observation into functional requirements $G_F$, a task-relevant perception vocabulary, and an inspection ranking. Scene observations incrementally update $G_O^k$, while semantic and geometric verification support joint object--role grounding to obtain $\phi^*$. Once the scene is functionally sufficient, the grounded assignment and observed state instantiate the symbolic planner, which produces the executable action sequence. The complete framework is described in Section~\ref{sec:methodology}.}
    \label{fig:overview}
    \vspace{-0.45cm}
\end{figure*}

To address the above challenges, we introduce \textsc{GRAB-TAMP} \textit{(Grounded Role Assignment Bridging Incomplete Scene Knowledge and Task and Motion Planning)}, a framework that bridges incomplete scene observations to downstream TAMP through functional grounding. \textsc{GRAB-TAMP} derives task-level functional requirements, incrementally acquires scene evidence when those requirements cannot yet be grounded, and jointly assigns verified physical entities to the required roles. An overview of the framework is shown in Fig.~\ref{fig:overview}.

The contributions of this work are threefold:
\begin{itemize}
    \item To our knowledge, this is the first foundation-model-guided TAMP framework to explicitly search for functionally required but currently unobserved entities and to use semantic and geometric verification to validate newly observed candidates.
    \item We formulate \emph{functional grounding} under incomplete scene knowledge as a joint assignment between task-level roles and physical scene entities, capturing object-level requirements together with relational, cardinality, sharing, reuse, and distinctness constraints.
    \item We introduce \emph{functional sufficiency} as an explicit criterion for determining when accumulated scene evidence is adequate for planning. Task planning proceeds only when all required functional roles admit a complete and jointly valid grounding to physical entities.

\end{itemize}
        
\section{Related Work}
\vspace{-0.1cm}
\subsection{Foundation-Model-Assisted Task and Motion Planning}
\vspace{-0.15cm}
Classical TAMP approaches require an interface between the task planner and the motion planner. PDDLStream~\cite{garrett2020pddlstream} combines symbolic planning with streams that sample continuous quantities such as grasps, placements, and trajectories. However, this requires having the task domain, including its predicates and actions, together with the relevant objects and their relations in the scene, to be specified beforehand.

Foundation Models (FMs) have recently reduced the amount of high-level domain structure that must be provided manually. Large Language Models (LLMs) have been used to construct the planning representation directly from natural-language instructions. AutoTAMP~\cite{chen2024autotamp} translates natural-language instructions into formal planning specifications, while CaStL~\cite{guo2025castl} extracts task requirements from language and translates them into planner-checkable constraints. ViLaIn-TAMP~\cite{siburian2025vilaintamp} uses vision-language reasoning to construct a formal task representation. In these approaches, the FM is used primarily to bridge the gap between a task description and the structured representation required by downstream planning.

Other work uses FMs to guide the TAMP search. LLM$^3$~\cite{wang2024llm3} reasons from motion-planning failures to revise task-level decisions. Rather than waiting for such failures, VLM-TAMP~\cite{yang2025vlmtamp} uses visually generated intermediate subgoals to reduce the effective planning horizon. ROBUST-TAMP~\cite{ojha2026planwayeventtriggeredfoundationmodel} performs event-triggered replanning as new objects are observed or execution conditions change. VIZ-COAST~\cite{yan2025vizcoast} reasons about constraints associated with downward-refinement failures, while OWL-TAMP~\cite{kumar2026owltamp} uses VLM-generated symbolic and continuous constraints to support open-world manipulation. Together, these works show that FMs can influence several aspects of TAMP, from specifying and decomposing the problem to constraining, adapting, and repairing the search.

\subsection{Planning with Incomplete Scene Knowledge}
\vspace{-0.15cm}
In TAMP, incomplete scene knowledge has largely been treated as uncertainty about the current state of the environment. Relevant objects may be hidden, their states may be unknown, and additional observations may therefore be required to determine whether a feasible plan can be synthesized.

Partial observability has been considered in TAMP through belief-space planning and explicit information-gathering actions~\cite{kaelbling2013IJRR}.
Similarly, TAMPURA~\cite{curtis2024tampura} models uncertainty over the initial state and action outcomes and plans over closed-loop controllers while reasoning about information-gathering actions and their associated risk. More recently, methods for TAMP under partial observability have combined task-level relevance reasoning with exploration so that objects required by the task can be progressively discovered~\cite{ma2025partiallyobservable}. These approaches generally assume that the task-relevant entities are already known, leaving the planner to reason about their location, state, or observability. In contrast, \textsc{GRAB-TAMP} first determines which physical scene entities can satisfy the functional roles required by the task, given RGB images and goal instruction.

FMs have also been used to reason about incomplete scene information. Zhao et al.~\cite{zhao2025seeing} construct a symbolic belief representation using FM uncertainty estimates and plan strategic information-gathering actions when task-relevant facts cannot be determined from the current observation. ROBUST-TAMP~\cite{ojha2026planwayeventtriggeredfoundationmodel} plans incrementally as task-relevant objects are discovered, but cannot handle tasks requiring complete object knowledge before planning. CoCo-TAMP~\cite{kim2026cocotamp} introduces LLM commonsense knowledge into hidden-state estimation for partially observable TAMP. VAP-TAMP~\cite{oloo2026vaptamp} employs active perception to gather additional information. These methods use FMs to reduce uncertainty or guide observation, but do not impose a pre-planning functional-sufficiency criterion based on complete joint role assignment. \textsc{GRAB-TAMP} instead evaluates whether observed entities are functionally sufficient before planning.
\subsection{Functional Grounding and Object--Role Assignment}
\vspace{-0.15cm}
Object identification alone is insufficient for task execution; an object must also satisfy the physical requirements of its intended use. Prior work addresses this gap through affordance learning, object property prediction, and geometric grounding, enabling task-relevant properties to be inferred directly from scene observations by estimating affordances for previously unseen objects~\cite{curtis2022affordances}, grounding symbolic object references to physical scene instances~\cite{ding2022groundobjects}, and predicting spatial affordances or interaction-relevant locations from visual observations~\cite{yuan2025robopoint}. However, these methods primarily ground individual objects, skills, predicates, or actions, whereas many tasks require jointly reasoning about multiple objects and their assignments. \textsc{GRAB-TAMP} addresses this gap by jointly grounding objects to task-level functional roles.

\section{Problem Formulation}
\vspace{-0.15cm}
\label{sec:problem_formulation}
\subsection{Task and Scene Representation}
\vspace{-0.15cm}
We consider an environment containing a finite set of physical entities $\C{O}$, including objects and placement regions and a finite set of inspectable regions $\C{R}$. The robot initially observes only part of the environment through observation $Z_0$. After a sequence of $k$ measurements ($Z_{0:k}$), the objects and the inspectable regions observed thus far are given by $\C{O}_k \subseteq \C{O}$, and $\C{I}_k \subseteq \C{R}$, respectively. Objects that have not yet been observed ($\C{O} \setminus \C{O}_k$) are treated as \emph{unknown}, rather than assumed to be absent. 

We denote by $Q=\{q_1,\ldots,q_m\}$ the set of functional roles required by the task. A role describes what an object must be capable of doing, rather than the identity of the object expected to do it. Roles with cardinality greater than one are expanded into indexed role instances in $Q$, while $\mathcal{A}_F$ specifies the corresponding sharing, reuse, and distinctness constraints. For an observed object $o \in \C{O}_k$, its compatibility with role $q_i$ is written as $
C_i(o)
=
C_i^{\mathrm{sem}}(o)
\land
C_i^{\mathrm{geom}}(o) $,
combining semantic compatibility with the geometric properties needed to perform that function. Some requirements cannot be checked one object at a time. We therefore represent relations between objects assigned to different roles using functional edges
\vspace{-0.2cm}
\begin{equation}
E_F\subseteq\mathcal{Q}\times\mathcal{Q}\times\mathcal{P}
\label{eq:functional-relations}
\vspace{-0.2cm}
\end{equation}
where each edge specifies a predicate in $\C{P}$ that must hold between the objects assigned to the corresponding pair of roles. For example, when $q_i \equiv \text{stirrer}, q_j \equiv \text{coffee-container}$, the corresponding compatibility predicate may evaluate to $1$, whereas for $q_j\equiv\text{soup-container}$, it may evaluate to $0$ (Fig.~\ref{fig:teaser}). Similarly, a task may also impose assignment constraints $\mathcal{A}_F$, for example, requiring two roles to use distinct objects or allowing the same object to be shared when the task permits it. The functional requirements are thus represented by
\vspace{-0.2cm}
\begin{equation}
G_F=(\mathcal{Q},E_F)
\label{eq:functional-graph}
\vspace{-0.25cm}
\end{equation}
\noindent together with the assignment constraints $\mathcal{A}_F$. Rather than specifying which particular objects should be used, we describe the task through the functions that objects need to perform. To this end, we define the task specification
\vspace{-0.2cm}
\begin{equation}
    \Psi = \left(G_F,\Gamma,\mathcal{V}_g, \mathcal{A}_F,\pi_R\right)
    \label{eq:task-specification}
    \vspace{-0.2cm}
\end{equation}
where $G_F$ captures the functional requirements of the task, $\Gamma$ specifies relations to achieve, while grounding checks functional compatibility, $\mathcal{V}_g$ provides a task-conditioned vocabulary for perception, and $\pi_R$ gives region inspection order. $\mathcal{V}_g$, $\pi_R$ guide what the system looks for and where it searches, but do not determine the final object--role assignment. We maintain an observed graph
\vspace{-0.2cm}
\begin{equation}
G_O^k=(\mathcal{O}_k,E_O^k)
\label{eq:observed-graph}
\vspace{-0.2cm}
\end{equation}
containing the objects and relations observed up to step $k$. The distinction between the two graphs is useful here: $G_F$ stays fixed and describes what the task needs, while $G_O^k$ grows as more of the environment becomes visible. Functional grounding therefore determines whether the objects observed so far collectively satisfy the requirements encoded by $G_F$.
\subsection{Functional Grounding}
\vspace{-0.15cm}
At inspection step $k$, a grounding is a mapping $
\phi_k:\mathcal{Q}\rightarrow\mathcal{O}_k
\label{eq:grounding-mapping}
$, which assigns an observed entity to each functional role. A valid grounding must satisfy the unary requirements of every role, 
\vspace{-0.2cm}
\begin{equation}
C_i\!\left(\phi_k(q_i)\right)=1,
\qquad
\forall q_i\in\mathcal{Q}
\label{eq:unary-requirements}
\end{equation}
as well as every relation encoded in $E_F$,
\vspace{-0.2cm}
\begin{equation}
p\!\left(\phi_k(q_i),\phi_k(q_j)\right)=1,
\qquad
\forall(q_i,q_j,p)\in E_F
\label{eq:relational-requirements}
\end{equation}
Assignments must also satisfy the sharing, distinctness, and set-level constraints in $\mathcal{A}_F$. We collect all such assignments in
\vspace{-0.2cm}
\begin{equation}
\Phi_k =
\left\{
\phi:\mathcal{Q}\rightarrow\mathcal{O}_k
\;\middle|\;
\substack{
C_i(\phi(q_i))=1,\ \forall q_i,\\
p(\phi(q_i),\phi(q_j))=1,\ \forall (q_i,q_j,p)\in E_F,\\
\phi \models \mathcal{A}_F
}
\right\}
\label{eq:valid-groundings}
\vspace{-0.1cm}
\end{equation}
The key aspect is that grounding is evaluated \emph{jointly} across the assigned objects. An object may satisfy the requirements of a role in isolation, yet the resulting assignment may be invalid when the required relations between roles are considered. For example, in Fig.~\ref{fig:teaser}, ``serving spoon 2" satisfies the serving role but is incompatible with the smaller soup bowl. Consequently, an FM-proposed candidate is not accepted solely because it is semantically preferred. The task-conditioned vocabulary $\mathcal{V}_g$ guides perception toward relevant objects, while membership in $\Phi_k$ is determined by the verified functional constraints above.
\begin{table}[t]
\centering
\caption{Components in task specification and functional grounding.}
\label{tab:grounding}
\setlength{\tabcolsep}{3pt}
\renewcommand{\arraystretch}{1.04}
\footnotesize
\vspace{-0.2cm}
\scalebox{0.9}{
\begin{tabular}{@{}c l@{}}
\toprule
\textbf{Component} & \textbf{Role in the framework} \\
\midrule
$\mathcal{V}_g$ & Specifies what perception should look for. \\
$\pi_R$         & Defines where the scene should be inspected next. \\
$G_F$           & Represents the functional requirements of the task. \\
$G_O^k$         & Stores the scene information observed up to step $k$. \\
$\phi$          & Maps functional roles to observed physical objects. \\
$\Phi_k$        & Contains the assignments that satisfy the functional requirements. \\
$\phi_k^*$      & Denotes the valid assignment selected for downstream planning. \\
\bottomrule
\end{tabular}}
\vspace{-0.45cm}
\end{table}
Whenever \(\Phi_k\neq\emptyset\), the grounding module selects a valid assignment \(\phi_k^*\in\Phi_k\), which is then fixed for downstream planning. The variables are summarized in Table~\ref{tab:grounding}. 

\vspace{-0.1cm}
\subsection{Functional Sufficiency, Search, and Planning}
\vspace{-0.1cm}
We say that the currently observed scene is \emph{functionally sufficient} when at least one complete valid grounding exists:
\vspace{-0.2cm}
\begin{equation}
\operatorname{Sufficient}(G_O^k,G_F)
=
\begin{cases}
1, & \Phi_k\neq\emptyset,\\
0, & \Phi_k=\emptyset
\end{cases}
\label{eq:functional-sufficiency}
\vspace{-0.2cm}
\end{equation}
This gives the search process a simple stopping condition. If $\Phi_k\neq\emptyset$, the currently observed objects are sufficient for functional grounding and no additional inspection is needed. If $\Phi_k=\emptyset$ while uninspected regions remain, the next region is selected according to $\pi_R$, inspected, and incorporated into $G_O^k$. Grounding is then reconsidered over the expanded observation.

If all regions in \(\mathcal{R}\) have been inspected and \(\Phi_k\) remains empty, the task is declared \emph{functionally infeasible} within the explored scene, since no assignment satisfying \(G_F\) could be found. Functional sufficiency, however, does not by itself guarantee that the manipulation task is executable. Once a valid assignment $\phi^{*}_k$ has been obtained, the abstract roles in $G_F$ are replaced by their corresponding physical objects and, together with $\Gamma$, accumulated scene information, and a fixed symbolic action domain $\C{D}$, are used to construct the downstream TAMP problem. If no plan exists despite having a valid functional grounding, the outcome is instead a \emph{planning failure}. This keeps two different questions separate: whether suitable objects can be found and assigned to the required functions, and whether those grounded objects admit a feasible manipulation plan.
\raggedbottom
\begin{algorithm}[t]
\caption{\textsc{GRAB-TAMP}}
\label{alg:functional_grounding_tamp}
\small
\textbf{Input:} Task goal $g$, initial observation $Z_0$, inspectable regions $\mathcal{R}$, symbolic domain $\mathcal{D}$. \\
\textbf{Output:} Executable task plan $\pi$, or $\textsc{Failure}$ with reasoning ($\Delta$).

\begin{algorithmic}[1]
\State $(G_F,\Gamma,\mathcal{V}_g, \mathcal{A}_F, \pi_R) \gets \textsc{FMPropose}(g,Z_0)$ \texttt{//} \textcolor{mygray}{Functional specification and search guidance}
\State $G_O^0 \gets \emptyset$; $\mathcal{I} \gets \emptyset$; $Z \gets Z_0$; $k \gets 1$
\texttt{//} \textcolor{mygray}{Persistent observed graph, Inspected regions, Initial observation}

\While{\textbf{true}}
    \State $G_O^k \gets \textsc{UpdateObsGraph}(G_O^{k-1}, Z, \mathcal{V}_g)$
    \texttt{//} \textcolor{mygray}{Add newly observed semantic and geometric evidence}  
    \State $\mathcal{C}^k \gets \textsc{VerifyCandidates}(G_O^k, G_F)$
    \texttt{//} \textcolor{mygray}{Semantic, unary-geometric, and relational verification} 
    \State $\Phi^k \gets \textsc{GroundRoles}(\mathcal{C}^k, G_F, \mathcal{A}_F)$
    \texttt{//} \textcolor{mygray}{Joint assignments satisfying roles, relations, and reuse constraints}  
    \If{$\Phi^k \neq \emptyset$}
        \State $\phi^* \gets \textsc{SelectGrounding}(\Phi^k)$
        \texttt{//} \textcolor{mygray}{Fix one verified complete assignment}  
        \State $P \gets \textsc{BuildSymbolicProblem}(\mathcal{D}, G_O^k, G_F, \phi^*, \Gamma)$
        \texttt{//} \textcolor{mygray}{Instantiate grounded task and scene state}  
        \State $\pi \gets \textsc{AStar}(P)$
        \texttt{//} \textcolor{mygray}{Synthesize manipulation sequence} 
        \If{$\pi \neq \textsc{Infeasible}$}
            \State \Return $\pi$
        \Else
            \State $\Delta \gets \textsc{PlanningFailureReason}\! (G_O^k,\! G_F,\! \phi^*,\! \Gamma)$
            \State \Return $(\textsc{PlanningFailure}, \Delta)$
        \EndIf
    \EndIf
    \If{$\mathcal{I} = \mathcal{R}$}
        \State $\Delta \gets \textsc{GroundingFailureReason}(G_O^k, G_F, \mathcal{C}^k)$
        \State \Return $(\textsc{GroundingFailure}, \Delta)$
    \EndIf
    \State $r \gets \textsc{NextRegion}(\pi_R, \mathcal{R}\setminus\mathcal{I})$
   \texttt{//} \textcolor{mygray}{Follow the FM-provided inspection ordering}  
    \State $Z \gets \textsc{InspectAndObserve}(r)$
    \texttt{//} \textcolor{mygray}{New scene information}  
    \State $\mathcal{I} \gets \mathcal{I} \cup \{r\}$
    \State $k \gets k + 1$
\EndWhile

\end{algorithmic}
\end{algorithm}
\flushbottom

\vspace{-0.2cm}

\vspace{-0.1cm}
\section{Methodology}
\vspace{-0.1cm}
\label{sec:methodology}
An overview of \textsc{GRAB-TAMP} is given in Algorithm~\ref{alg:functional_grounding_tamp}. The framework builds the observed graph incrementally and evaluates the available objects against the fixed functional requirements. The following subsections describe how the functional graph \(G_F\) is generated, how \(G_O^k\) is updated, how candidates are verified and jointly grounded, and how the frozen grounding is compiled into the downstream planner.

\vspace{-0.15cm}
\subsection{FM Functional Specification and Search Guidance}
\vspace{-0.15cm}
Given the task instruction $g$ and initial observation $Z_0$, the FM provides the task specification $\Psi$ (line 1). Rather than directly selecting the physical objects to be used, the FM first specifies the task-level roles and the properties and relations that must be verified for each role. The resulting specification is mapped onto the nodes and edges of $G_F$, together with the assignment constraints in $\mathcal{A}_F$, when the task requires objects to remain distinct, to be reused, or to be shared across roles. Candidate categories proposed for each role are collected into \(V_g\) and used only to guide open-vocabulary perception. They do not determine the final grounding. Whether an observed entity satisfies a role is decided later through the verification and joint assignment stages. The FM also provides the region ordering $\pi_R$, which determines the order in which currently uninspected regions are considered when additional scene information is required. In this way, the FM contributes to both the functional specification and the inspection guidance, where relevant objects may be found, as shown in Fig.~\ref{fig:overview}.

\vspace{-0.15cm}
\subsection{Incremental Observed-Graph Construction}
\vspace{-0.125cm}
The initial observation \(Z_0\) is processed first by the observed graph \(G_O^0\). Objects detected in the visible scene are added to $G_O^0$ together with their associated region and the available semantic and geometric evidence (line 4). The next uninspected region is selected by subroutine \textsc{NextRegion} according to \(\pi_R\) (line 22), after which \textsc{InspectAndObserve} acquires the new observation (line 23). The resulting evidence is incorporated into \(G_O^k\) through subroutine \textsc{UpdateObsGraph}. Inspection of this region yields a new observation, from which newly visible objects and associated evidence are incorporated into \(G_O^k\). Object attributes and relations can be extended as new evidence becomes available, but objects that have already been verified remain part of the graph after each inspection update. 

\vspace{-0.15cm}
\subsection{Semantic and Geometric Verification}
\vspace{-0.15cm}
Candidate verification is performed by \textsc{VerifyCandidates} (line 5, Algorithm~1), which evaluates the semantic, unary geometric, and required pairwise checks described below. We use YOLO-World~\cite{cheng2024yoloworld} with the task-conditioned vocabulary $\mathcal{V}_g$ to identify observed objects that are plausible candidates for each functional role. Objects that do not satisfy the corresponding semantic requirement $C_i^{\mathrm{sem}}$ are discarded for that role.
The remaining candidates are then evaluated geometrically. For each object, its segmented depth observation is converted into object-level geometric evidence from which the required unary properties are evaluated. These checks instantiate $C_i^{\mathrm{geom}}$ for the corresponding role and include task-dependent properties such as usable dimensions, cavity structure, support geometry, or tool shape.
Required pairwise constraints are then evaluated according to~\eqref{eq:relational-requirements}. For each edge \((q_i,q_j,p)\in E_F\), the predicate \(p\) is evaluated on candidate entity pairs associated with roles \(q_i\) and \(q_j\), retaining only pairs for which $p = 1$. This avoids constructing every possible geometric relation between every pair of objects in the scene. Thus, verification progressively narrows detected objects to semantically compatible candidates satisfying unary geometric requirements, before using pairwise relations to identify complete grounding.

\vspace{-0.1cm}
\begin{figure}[t]
\centering
\includegraphics[
width=0.85\linewidth,
trim={0.68cm 19.1cm 12.2cm 0.8cm},
clip
]{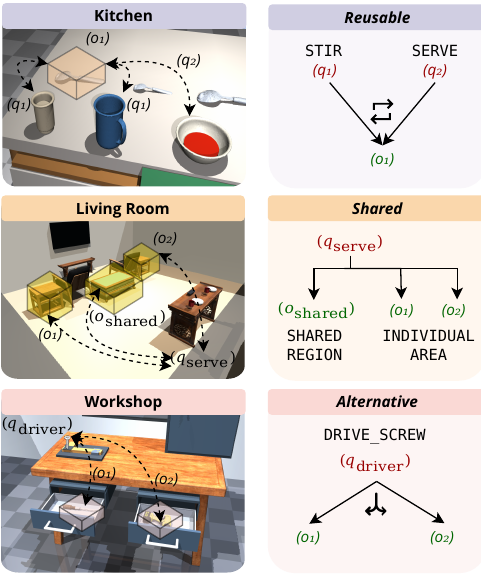}
\vspace{-0.18cm}
\caption{Functional assignments need not be one-to-one. Objects may be reused, regions may be shared, and multiple candidates may satisfy the same task role, as formulated by $\mathcal{A}_F$ in~\eqref{eq:valid-groundings}.}
\label{fig:3}
\vspace{-0.55cm}
\end{figure}

\begin{table*}[t]
    \centering
    \caption{Task and scene definitions used for evaluating functional grounding.}
    \label{tab:task_definitions}

    \vspace{-0.20cm}

    \fontsize{7.5}{8.1}\selectfont
    \setlength{\tabcolsep}{1.8pt}
    \renewcommand{\arraystretch}{1.00}

    \begin{tabularx}{\textwidth}{
        @{}
        >{\raggedright\arraybackslash}p{0.075\textwidth}
        >{\raggedright\arraybackslash}X
        >{\raggedright\arraybackslash}p{0.150\textwidth}
        >{\raggedright\arraybackslash}p{0.160\textwidth}
        >{\raggedright\arraybackslash}p{0.145\textwidth}
        >{\centering\arraybackslash}p{0.055\textwidth}
        @{}
    }

        \toprule

        \textbf{Domain} &
        \textbf{Goal instruction} &
        \textbf{Functional relations} &
        \textbf{Hidden objects} &
        \textbf{Candidate regions} &
        \makecell{\textbf{Variants}\\[-1pt]\textbf{F / I}} \\

        \midrule


        \textbf{\textit{Kitchen}}
        &
        Prepare and serve one coffee and one soup for each of two people.
        Make each coffee using coffee and water and stir it before serving.
        Serve each soup bowl with its own suitable eating utensil.
        &
        \texttt{OPEN\_CAVITY},
        \texttt{ELONGATED\_OBJECT},
        \texttt{INSERTABLE\_IN}, and
        \texttt{REACHES\_BOTTOM}.
        &
        Cups, bowls, stirring and eating utensils; required objects may
        be distributed across storage regions.
        &
        Countertop, drawers \texttt{D1--D2}, cabinets
        \texttt{C1--C2}, and bin \texttt{B1}.
        &
        6 / 6
        \\

        \addlinespace[1.5pt]


        \textbf{\textit{Living Room}}
        &
        Prepare the living room for two people to enjoy refreshments while
        watching television. Provide each person with their own refreshment
        setting nearby, and place the entertainment control where it is
        accessible to both people.
        &
        \texttt{PLANAR\_SUPPORT},
        \texttt{FITS\_ON},
        \texttt{FITS\_SET\_ON}, and
        \texttt{NEAR\_SEAT}.
        &
        None; task objects and placement regions are initially visible.
        &
        Left and right personal side tables and the shared coffee table.
        &
        6 / 4
        \\

        \addlinespace[1.5pt]


        \textbf{\textit{Workshop}}
        &
        Identify the compatible components required to complete the fastening
        at the marked workbench location, complete the fastening, and leave
        any reusable equipment used for the task safely on the workbench.
        &
        \texttt{GRASPABLE},
        \texttt{FITS\_TOOL\_HEAD},
        \texttt{FITS\_IN}, and
        \texttt{NEAR\_WORKPIECE}.
        &
        Manual and power drivers, compatible Phillips screw, and
        wooden-hammer distractor.
        &
        Left drawer, right drawer, tool cabinet, and marked workbench target.
        &
        8 / 2
        \\

        \bottomrule

    \end{tabularx}

    \vspace{1pt}

    \begin{minipage}{0.985\textwidth}
        \scriptsize
        \textit{Note:} F / I denotes feasible / infeasible variants.
    \end{minipage}

    \vspace{-0.4cm}
\end{table*}

\subsection{Joint Object--Role Assignment}
\vspace{-0.1cm}
The verified candidates are passed to \textsc{GroundRoles} (line 6), which constructs the set of complete assignments \(\Phi_k\) satisfying the role, relational, and assignment constraints. Candidate objects are not selected independently for each role, but rather together so that the required relations in $E_F$ and the assignment constraints in $\mathcal{A}_F$ are satisfied by the same mapping. This matters when multiple individually valid candidates exist, since satisfying unary role constraints does not guarantee compatibility with other role assignments. Fig.~\ref{fig:3} illustrates how the same grounding formulation accommodates reusable, shared, and alternative assignments across the three task domains. Distinct roles are prevented from taking the same physical instance, while roles that permit reuse may refer to an already assigned object when the remaining functional constraints are also satisfied. Shared assets are treated similarly, allowing one physical object to support multiple compatible parts of the task without requiring a duplicate instance.

If the current $G_O^k$ does not provide any valid complete assignment, control returns to the inspection stage and the graph is expanded using the next region in $\pi_R$. Once \(\Phi_k\neq\emptyset\), \textsc{SelectGrounding} selects one verified mapping \(\phi^*\) for downstream planning (line 8). At this point the object--role mapping is fixed. If no complete grounding exists after all regions are inspected, \textsc{GroundingFailureReason} reports the unresolved grounding conditions (line 19).
\vspace{-0.2cm}
\subsection{Grounded Symbolic Planning and Execution}
\vspace{-0.15cm}
The selected grounding $\phi^{}$ provides the mapping between functional grounding and downstream planning, and replaces the abstract roles appearing in the task specification with the corresponding physical objects selected by $\phi^{}$. The accumulated scene information in $G_O^k$ supplies the relevant object and region state, while $\Gamma$ specifies the task-level conditions that the resulting plan must achieve.
These inputs are passed to \textsc{BuildSymbolicProblem} (line 9), which constructs the grounded symbolic planning problem with action predicates. An A$^\star$ search is then performed over this state space to obtain an ordered manipulation sequence: $
\pi = [a_1,a_2,\ldots,a_T]$.

The planner operates only after the functional roles have been resolved to physical objects. Scene inspection is therefore not included as a symbolic planning action, and the planner does not modify the frozen grounding $\phi^{*}$. If no symbolic plan can be found after functional grounding, this is treated separately from the earlier case where no valid object-role assignment can be established from the explored scene. The resulting action sequence is passed to the execution layer, where each symbolic action is executed by the motion planner. If no symbolic plan is found despite a valid grounding, \textsc{PlanningFailureReason} identifies the downstream planning failure (line 14), keeping this case distinct from grounding failure. This preserves a clear boundary between scene search and functional grounding on one side, and task planning and physical execution on the other.

\vspace{-0.2cm}

\vspace{-0.2cm}
\section{Experimentation}
\vspace{-0.1cm}
\subsection{Environment Variants and Setup}
\vspace{-0.15cm}
All experiments are conducted in MuJoCo using the Google Robot mobile manipulator. We evaluate across three task domains---\textit{Kitchen}, \textit{Living Room}, and \textit{Workshop}, comprising 32 scene variants in total: 20 feasible and 12 infeasible, each evaluated over 10 trials~(see Table~\ref{tab:task_definitions}). \textit{Kitchen} includes hidden objects distributed across multiple storage regions, with varying cardinality and object reuse requirements; \textit{Living Room} includes multiple placement and shared-region constraints with fully observable objects; and \textit{Workshop} has alternative tools and hidden distractor objects. Each scene is observed through a fixed three-view camera configuration. We conduct 10 trials per scene variant, resulting in 320 trials for ablation and each baseline comparison. All inference calls use Qwen3.5--9B with full unquantized weights, served through vLLM on an inference server equipped with an NVIDIA~RTX~PRO~5000 Blackwell GPU card. All methods share the same budget, at most 15 model calls per trial, a 24,576-token generation ceiling and the checkpoint's published sampling configuration beyond which we observed diminishing and saturated returns for the average of all baseline comparisons, including our work.

\begin{table}[t]
\centering
\caption{End-to-end comparison of 3 task domains using Google Robot. Feasible and infeasible variants are
reported separately.}
\label{tab:main_results}
\vspace{-0.2cm}
\setlength{\tabcolsep}{1.25pt}
\renewcommand{\arraystretch}{1.03}
\scriptsize

\begin{tabular*}{\columnwidth}{
@{\extracolsep{\fill}}
l
ccc
c
cc
@{}
}
\toprule

& \multicolumn{3}{c}{\textbf{Feasible}}
& \multicolumn{1}{c}{\textbf{All trials}}
& \multicolumn{2}{c}{\textbf{Infeasible}} \\

\cmidrule(lr){2-4}
\cmidrule(lr){5-5}
\cmidrule(l){6-7}

\textbf{Method} &
\textbf{$N_F$} &
\makecell{\textbf{PGC}\\ } &
\makecell{\textbf{E2E}\\\textbf{succ.}} &
\makecell{\textbf{Avg.}\\\textbf{replans}} &
\textbf{$N_I$} &
\makecell{\textbf{NC}\\} \\

\midrule


\multicolumn{7}{@{}l}{
    \textbf{\textit{Kitchen}}
    \hfill
    \footnotesize\textit{hidden across five storage regions}
} \\
\addlinespace[1pt]

VLM-TAMP~\cite{yang2025vlmtamp}
    & 60 & 67.5 & 0.0
    & 12.6
    & 60 & 0.0 \\

OWL-TAMP~\cite{kumar2026owltamp}
    & 60 & 45.5 & 0.0
    & 0.0
    & 60 & 13.3 \\

ROBUST-TAMP~\cite{ojha2026planwayeventtriggeredfoundationmodel}
    & 60 & 49.2 & 0.0
    & 11.9
    & 60 & 25.0 \\

\textbf{GRAB-TAMP}
    & 60
    & \textbf{82.8}
    & \textbf{65.0}
    & \textit{--}
    & 60
    & \textbf{50.0} \\

\midrule


\multicolumn{7}{@{}l}{
    \textbf{\textit{Living Room}}
    \hfill
    \footnotesize\textit{all objects initially visible}
} \\
\addlinespace[1pt]

VLM-TAMP~\cite{yang2025vlmtamp}
    & 60 & \textbf{100.0} & 80.0
    & 7.5
    & 40 & 0.0 \\

OWL-TAMP~\cite{kumar2026owltamp}
    & 60 & \textbf{100.0} & 60.0
    & 0.0
    & 40 & 5.0 \\

ROBUST-TAMP~\cite{ojha2026planwayeventtriggeredfoundationmodel}
    & 60 & \textbf{100.0} & \textbf{88.3}
    & 6.2
    & 40 & 0.0 \\

\textbf{GRAB-TAMP}
    & 60
    & 48.7
    & 31.7
    & \textit{--}
    & 40
    & \textbf{70.0} \\

\midrule


\multicolumn{7}{@{}l}{
    \textbf{\textit{Workshop}}
    \hfill
    \footnotesize\textit{hidden alternative candidates}
} \\
\addlinespace[1pt]

VLM-TAMP~\cite{yang2025vlmtamp}
    & 80 & 99.6 & 26.2
    & 11.9
    & 20 & 0.0 \\

OWL-TAMP~\cite{kumar2026owltamp}
    & 80 & \textbf{100.0} & 15.0
    & 2.62
    & 20 & 0.0 \\

ROBUST-TAMP~\cite{ojha2026planwayeventtriggeredfoundationmodel}
    & 80 & 17.5 & 0.0
    & 7.5
    & 20 & \textbf{70.0} \\

\textbf{GRAB-TAMP}
    & 80
    & 69.6
    & \textbf{62.5}
    & \textit{--}
    & 20
    & 20.0 \\


\specialrule{0.7pt}{2.5pt}{1.5pt}

\textbf{GRAB-TAMP (Overall)}
    & \textbf{200}
    & \textbf{67.3}
    & \textbf{54.0}
    & \textbf{\textit{--}}
    & \textbf{120}
    & \textbf{51.7} \\

\bottomrule
\end{tabular*}

\vspace{2pt}

\begin{minipage}{0.98\columnwidth}
\scriptsize
\textit{Note:}
\textbf{PGC}, \textbf{E2E success}, and \textbf{NC} are reported in \%.
\textbf{Avg. replans} is the mean number of planning re-invocations after the first, over all $N_F{+}N_I$ trials of a scene.
\end{minipage}
\vspace{-0.2cm}
\end{table}

\vspace{-0.1cm}
\subsection{Evaluation Metrics}
\vspace{-0.1cm}
Feasible and infeasible variants are evaluated separately because they measure different behaviors. Let $N_F$ and $N_I$ denote the number of feasible and infeasible trials, respectively. For trial $i$, let $G_i$ denote the set of required task goals and $\hat{G}_i$ the subset achieved
by the generated plan. We define:

\noindent \textbf{Plan Goal Coverage (PGC):} For feasible tasks, we measure fraction of required task goals achieved by the generated plan
\vspace{-0.2cm}
\begin{equation}
\mathrm{PGC}
=
\frac{1}{N_F}
\sum_{i=1}^{N_F}
\frac{|G_i \cap \hat{G}_i|}{|G_i|}
\times 100
\vspace{-0.2cm}
\end{equation}

\noindent \textbf{End-to-End (E2E) Success:}
Let $s_i \in \{0,1\}$ indicate whether feasible trial $i$ successfully
satisfies all task goals after execution.\\
\vspace{-0.6cm}
\begin{equation}
\mathrm{E2E \,\,Success}
=
\frac{1}{N_F}
\sum_{i=1}^{N_F}
s_i
\times 100
\vspace{-0.2cm}
\end{equation}
\noindent \textbf{Non-Commitment Rate (NC):}
Fraction of infeasible trials on which no plan is committed. It does not distinguish explicit infeasibility recognition from exhausted search or absent output,
\vspace{-0.4cm}
\begin{equation}
\mathrm{NC}
=
\frac{1}{N_I}
\sum_{i=1}^{N_I}
(1-C_i)
\times 100
=
100-\mathrm{Commit}
\vspace{-0.2cm}
\end{equation}
\vspace{-0.2cm}
\subsection{Overall Performance}
\vspace{-0.1cm}
Table~\ref{tab:main_results} compares GRAB-TAMP against three FM-assisted TAMP approaches under the same execution settings. Across the 200 feasible trials, GRAB-TAMP achieves 54.0\% E2E success, compared with 34.5\% for VLM-TAMP. VLM-TAMP attains higher PGC (90.1\% versus 67.3\%) and its substantially larger PGC-to-E2E gap indicates that partial plan coverage often does not translate to complete execution. The difference between PGC and complete success is important in this setting, since several methods can produce partial plans of a long-horizon task without resolving all requirements needed for complete execution.

The baseline methods differ mainly in where FM reasoning enters the planning pipeline and how TAMP is organized around it. VLM-TAMP achieves substantial PGC in \textit{Kitchen} (67.5\%) and \textit{Workshop} (99.6\%), but only translates to 0\% and 26.2\% E2E success, respectively. Its intermediate-subgoal guidance can therefore support partial task progress once suitable objects are available, but does not explicitly determine whether all hidden or geometrically constrained objects required for complete execution have been resolved before planning. With replanning enabled, OWL-TAMP achieves 100.0\% PGC and 15.0\% E2E success in Workshop, averaging 2.62 replans across all trials. Despite complete plan goal coverage, execution success remains limited. In \textit{Kitchen}, one planning cycle consumes the entire 15 call budget, leaving no room for a second and reducing the protocol to single-shot. ROBUST-TAMP performs best among the baselines in \textit{Living Room}, reaching 88.3\% E2E, but its performance falls sharply in \textit{Kitchen} and \textit{Workshop}. Although its event-triggered replanning allows the system to react to newly observed objects and execution changes, the resulting 49.2\% and 17.5\% PGC in these two domains indicate that reactive replanning alone does not resolve joint functional compatibility required when several hidden candidates must be considered together. 

On the other hand, \textsc{GRAB-TAMP} shows its largest gains in \textit{Kitchen} and \textit{Workshop}, achieving \(65.0\%\) and \(62.5\%\) E2E success, respectively, where task completion depends on discovering hidden objects or geometrically constrained functional candidates.
Its \(31.7\%\) success in \textit{Living Room} is lower than all three baselines, suggesting that grounding ambiguities can arise when relatively simple, fully observable tasks are decomposed into a richer set of functional roles than is necessary for execution. 
Across the 120 infeasible trials, GRAB-TAMP obtains a 51.7\% NC rate, compared with 24.2\% and 0.0\% for ROBUST-TAMP and VLM-TAMP, respectively. These rates do not always reflect explicit infeasibility recognition. For the baselines, any trial producing no executable plan counts as a noncommitment, conflating exhaustion with abstention. ROBUST-TAMP's 70.0\% in \textit{Workshop} is due to replanning budget exhaustion without emitting an action sequence; OWL-TAMP records 0.0\% NC in \textit{Workshop}, committing to a partial action sequence in every infeasible trial before the budget is exhausted, and appears better suited to shorter-horizon tasks under a fixed model-call budget. \textsc{GRAB-TAMP} instead retains the source of failure across grounding and planning, distinguishing an unresolved functional assignment from a downstream planning failure. 
\begin{table}[t]
\centering
\caption{Verification and inspection ablations.}
\label{tab:ablation}

\vspace{-0.20cm}

\setlength{\tabcolsep}{1.35pt}
\renewcommand{\arraystretch}{0.98}
\scriptsize


\begin{tabularx}{\columnwidth}{
@{}
>{\raggedright\arraybackslash}p{0.120\columnwidth}
>{\raggedright\arraybackslash}p{0.190\columnwidth}
>{\centering\arraybackslash}p{0.075\columnwidth}
>{\centering\arraybackslash}X
>{\centering\arraybackslash}X
>{\centering\arraybackslash}X
@{}
}

\toprule
\multicolumn{6}{c}{\textbf{Verification Ablation}} \\
\midrule

\textbf{Scene} &
\textbf{Verification} &
\textbf{Trials} &
\makecell[c]{\textbf{Matching}\\$\mathbf{G_O}$ (\%)} &
\makecell[c]{\textbf{Mismatching}\\$\mathbf{G_O}$ (\%)} &
\makecell[c]{\textbf{Missing}\\$\mathbf{G_O}$ (\%)} \\

\midrule

\multirow{3}{*}{\textit{Kitchen}}
& Semantic only & 60 & 96 (12.3) & 19 (2.4) & 665 (85.3) \\
& + Unary       & 60 & 103 (13.2) & 21 (2.7) & 656 (84.1) \\
& \textbf{+ Binary}
& 60 & \textbf{514 (65.9)} & 63 (8.1) & \textbf{203 (26.0)} \\

\addlinespace[1pt]

\multirow{3}{*}{\makecell[l]{\textit{Living}\\[-1pt]\textit{Room}}}
& Semantic only & 60 & 141 (18.1) & 1 (0.1) & 638 (81.8) \\
& + Unary       & 60 & 96 (12.3) & 3 (0.4) & 681 (87.3) \\
& \textbf{+ Binary}
& 60 & \textbf{439 (56.3)} & 20 (2.6) & \textbf{321 (41.2)} \\

\addlinespace[1pt]

\multirow{3}{*}{\textit{Workshop}}
& Semantic only & 80 & 86 (17.9) & 13 (2.7) & 381 (79.4) \\
& + Unary       & 80 & 47 (9.8) & 8 (1.7) & 425 (88.5) \\
& \textbf{+ Binary}
& 80 & \textbf{239 (49.8)} & 55 (11.5) & \textbf{186 (38.8)} \\

\specialrule{0.55pt}{2pt}{1.5pt}

\multirow{3}{*}{\textbf{Overall}}
& Semantic only & 200 & 323 (15.8) & 33 (1.6) & 1684 (82.5) \\
& + Unary       & 200 & 246 (12.1) & 32 (1.6) & 1762 (86.4) \\
& \textbf{+ Binary}
& 200 & \textbf{1192 (58.4)} & 138 (6.8) & \textbf{710 (34.8)} \\

\bottomrule
\end{tabularx}


\vspace{0.30em}

\begin{tabularx}{\columnwidth}{
@{}
>{\raggedright\arraybackslash}p{0.145\columnwidth}
>{\raggedright\arraybackslash}p{0.195\columnwidth}
>{\centering\arraybackslash}X
>{\centering\arraybackslash}X
>{\centering\arraybackslash}X
@{}
}

\multicolumn{5}{c}{\textbf{Inspection-Order Ablation}} \\
\midrule

\textbf{Scene} &
\textbf{Order} &
\textbf{Regions} &
\textbf{Checks} &
\textbf{Time (s)} \\

\midrule

\multirow{2}{*}{\textit{Kitchen}}
& Fixed order
& 4.33
& 134.6
& 82.94 \\

& \textbf{\textit{FM-ranked}}
& \textbf{3.91}
& 134.6
& \textbf{76.22} \\

\addlinespace[1pt]

\multirow{2}{*}{\makecell[l]{\textit{Living}\\[-1pt]\textit{Room}}}
& Fixed order
& 0.00
& 23.2
& 6.77 \\

& \textbf{\textit{FM-ranked}}
& 0.00
& 23.2
& 6.77 \\

\addlinespace[1pt]

\multirow{2}{*}{\textit{Workshop}}
& Fixed order
& 2.58
& 14.2
& 287.64 \\

& \textbf{\textit{FM-ranked}}
& \textbf{2.32}
& \textbf{13.4}
& \textbf{268.20} \\

\bottomrule
\end{tabularx}

\vspace{0.1em}

\begin{minipage}{0.985\columnwidth}
\fontsize{6.8}{7.2}\selectfont
\raggedright
\end{minipage}

\vspace{-0.5cm}
\end{table}

\vspace{-0.175cm}
\subsection{Ablation}
\vspace{-0.1cm}
We perform two ablations as shown in Table~\ref{tab:ablation}. First, we evaluate how semantic, unary geometric, and binary relational checks affect the resulting functional grounding. Second, we evaluate whether using the FM-derived region ranking improves the efficiency of scene inspection relative to a fixed inspection order. Semantic verification uses YOLO-World with the FM-derived vocabulary, unary verification adds object geometry, and binary verification checks relations between candidates. For example, a spoon is useful only if it fits or reaches the container bottom. These checks determine which nodes and relations are retained in $G_O^k$ without changing the joint-assignment procedure. Semantic and unary verification provide only $15.8\%$ and $12.1\%$ correct grounding evidence, while binary relations increase this to $58.4\%$ and reduce missing evidence from $82.5\%$ to $34.8\%$. Unary geometry applies a stricter filter and may reject semantically plausible candidates under noisy or incomplete geometric evidence, explaining the lower coverage before binary verification substantially improves joint assignment coverage. We also compare the FM-derived ranking $\pi_R$ with a fixed order, keeping the remaining pipeline unchanged. In \textit{Kitchen}, FM ranking reduces inspected regions from $4.33$ to $3.91$ and runtime from $82.94$\,s to $76.22$\,s; in \textit{Workshop}, the corresponding values decrease from $2.58$ to $2.32$ regions, $14.2$ to $13.4$ candidate checks, and from $287.64$\,s to $268.20$\,s. \textit{Living Room} is unchanged since all task-relevant entities are initially visible. Thus, FM ranking does not modify the grounding criterion, but reduces unnecessary scene inspection and reaches functional sufficiency with less search effort.

\vspace{-0.1cm}

\begin{figure}[tbp]
    \centering
    \includegraphics[
        width=0.75\linewidth,
        trim={1.4cm 1cm 2.2cm 0cm},
        clip
    ]{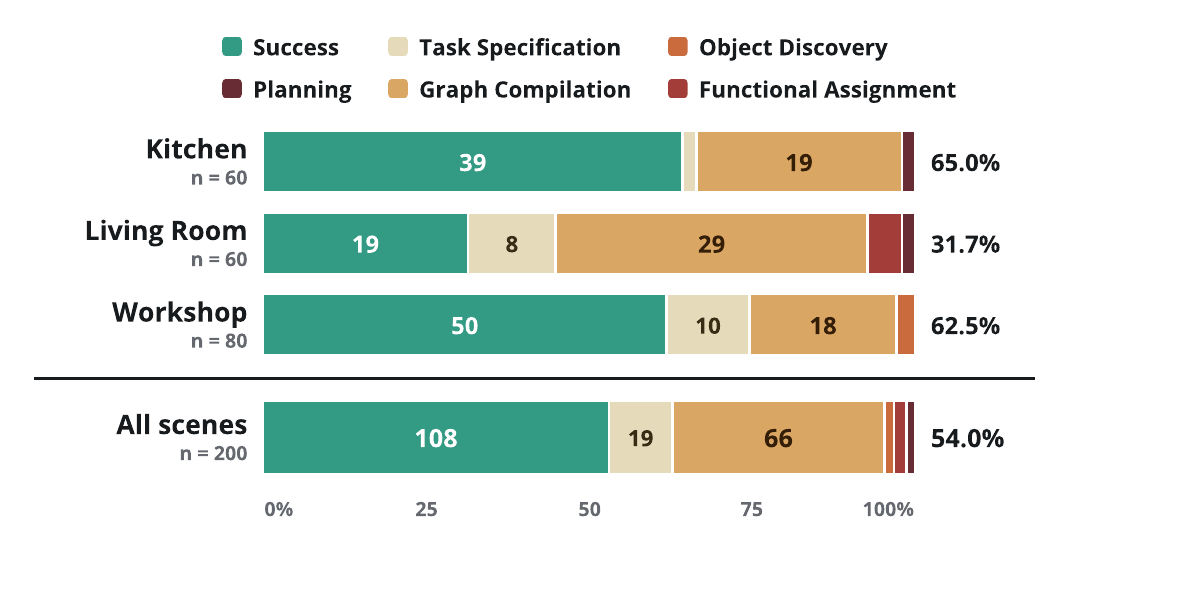}
      \caption{Failure breakdown of GRAB-TAMP for feasible variants. Categories give the earliest stage that failed, not the point of termination: only Task Specification precludes a plan (PGC = 0), while others retain partial plan and contribute nonzero goal coverage.}
    \label{fig:4}
    \vspace{-0.6cm}
\end{figure}

\section{Conclusion and Discussion}
\vspace{-0.1cm}
We propose functional sufficiency as an explicit criterion for determining when the available scene evidence is adequate for downstream TAMP, and use functional grounding in GRAB-TAMP to establish this condition before planning while distinguishing unresolved grounding from downstream planning failure. The ablations further show that semantic verification alone is insufficient for functional grounding, while geometric and relational verification improve assignment validity and FM-ranked inspection reduces unnecessary search and inspection time. Failures can still arise when the FM omits or misstates a required role, relation, cardinality, or binding constraint, or when perception provides incorrect geometric evidence, as shown in the figure breakdown in Fig.~\ref{fig:4}. Misstated roles may remain unresolved; omitted requirements introduce no goals into \(\Gamma\). From baseline comparisons, we find that end-to-end failure does not necessarily imply failed functional grounding, since partial task progress may still be achieved before all task requirements are satisfied. A practical limitation is that equivalent functional relations may be expressed differently across FMs and prompts, while the execution layer requires a fixed predicate vocabulary. We address this by mapping free-form relation descriptions to corresponding executable property checks, by using a zero-shot NLI DeBERTa~\cite{laurer2023building} model. Further, a functionally valid assignment does not guarantee downstream motion feasibility. Planning or execution may still fail because of reachability or other manipulation constraints, and geometric verification is limited by the accuracy of the estimated scene geometry.

\vspace{-0.15cm}

\bibliographystyle{IEEEtran}
\bibliography{references}

@inproceedings{garrett2020pddlstream,
  author    = {Caelan Reed Garrett and Tom{\'a}s Lozano-P{\'e}rez and Leslie Pack Kaelbling},
  title     = {{PDDLStream}: Integrating Symbolic Planners and Blackbox Samplers via Optimistic Adaptive Planning},
  booktitle = {Proceedings of the International Conference on Automated Planning and Scheduling (ICAPS)},
  volume    = {30},
  pages     = {440--448},
  year      = {2020}
}

@article{zhao2025seeing,
  title   = {Seeing is Believing: Belief-Space Planning with Foundation Models as Uncertainty Estimators},
  author  = {Zhao, Linfeng and McClinton, Willie and Curtis, Aidan and Kumar, Nishanth and Silver, Tom and Kaelbling, Leslie Pack and Wong, Lawson L. S.},
  journal = {arXiv preprint arXiv:2504.03245},
  year    = {2025},
  doi     = {10.48550/arXiv.2504.03245}
}

@article{garrett2021integrated,
  title   = {Integrated Task and Motion Planning},
  author  = {Garrett, Caelan Reed and Chitnis, Rohan and Holladay, Rachel and
             Kim, Beomjoon and Silver, Tom and Kaelbling, Leslie Pack and
             Lozano-P{\'e}rez, Tom{\'a}s},
  journal = {Annual Review of Control, Robotics, and Autonomous Systems},
  volume  = {4},
  pages   = {265--293},
  year    = {2021},
  doi     = {10.1146/annurev-control-091420-084139}
}

@inproceedings{huang2023inner,
  title     = {Inner Monologue: Embodied Reasoning through Planning with Language Models},
  author    = {Huang, Wenlong and Xia, Fei and Xiao, Ted and Chan, Harris and
               Liang, Jacky and Florence, Pete and Zeng, Andy and
               Tompson, Jonathan and Mordatch, Igor and Chebotar, Yevgen and
               Sermanet, Pierre and Jackson, Tomas and Brown, Noah and
               Luu, Linda and Levine, Sergey and Hausman, Karol and
               Ichter, Brian},
  booktitle = {Proceedings of the 6th Conference on Robot Learning},
  series    = {Proceedings of Machine Learning Research},
  volume    = {205},
  pages     = {1769--1782},
  year      = {2023},
  publisher = {PMLR}
}

@inproceedings{chen2024autotamp,
  author    = {Yongchao Chen and Jacob Arkin and Charles Dawson and Yang Zhang and Nicholas Roy and Chuchu Fan},
  title     = {{AutoTAMP}: Autoregressive Task and Motion Planning with {LLM}s as Translators and Checkers},
  booktitle = {2024 IEEE International Conference on Robotics and Automation (ICRA)},
  pages     = {6695--6702},
  year      = {2024}
}

@inproceedings{guo2025castl,
  author    = {Weihang Guo and Zachary Kingston and Lydia E. Kavraki},
  title     = {{CaStL}: Constraints as Specifications through {LLM} Translation for Long-Horizon Task and Motion Planning},
  booktitle = {2025 IEEE International Conference on Robotics and Automation (ICRA)},
  pages     = {11957--11964},
  year      = {2025}
}

@article{siburian2025vilaintamp,
  author  = {Jeremy Siburian and Keisuke Shirai and Cristian C. Beltran-Hernandez and Masashi Hamaya and Michael G{\"o}rner and Atsushi Hashimoto},
  title   = {Grounded Vision-Language Interpreter for Integrated Task and Motion Planning},
  journal = {arXiv preprint arXiv:2506.03270},
  year    = {2025}
}

@inproceedings{wang2024llm3,
  author    = {Shu Wang and Muzhi Han and Ziyuan Jiao and Zeyu Zhang and Ying Nian Wu and Song-Chun Zhu and Hangxin Liu},
  title     = {{LLM}$^3$: Large Language Model-Based Task and Motion Planning with Motion Failure Reasoning},
  booktitle = {2024 IEEE/RSJ International Conference on Intelligent Robots and Systems (IROS)},
  pages     = {12086--12092},
  year      = {2024}
}

@inproceedings{yang2025vlmtamp,
  author    = {Zhutian Yang and Caelan Garrett and Dieter Fox and Tom{\'a}s Lozano-P{\'e}rez and Leslie Pack Kaelbling},
  title     = {Guiding Long-Horizon Task and Motion Planning with Vision Language Models},
  booktitle = {2025 IEEE International Conference on Robotics and Automation (ICRA)},
  pages     = {16847--16853},
  year      = {2025}
}

@article{yan2025vizcoast,
  author  = {Muyang Yan and Miras Mengdibayev and Ardon Floros and Weihang Guo and Lydia E. Kavraki and Zachary Kingston},
  title   = {Using {VLM} Reasoning to Constrain Task and Motion Planning},
  journal = {arXiv preprint arXiv:2510.25548},
  year    = {2025}
}

@article{kumar2026owltamp,
  author  = {Nishanth Kumar and William Shen and Fabio Ramos and Dieter Fox and Tom{\'a}s Lozano-P{\'e}rez and Leslie Pack Kaelbling and Caelan Reed Garrett},
  title   = {Open-World Task and Motion Planning via Vision-Language Model Generated Constraints},
  journal = {IEEE Robotics and Automation Letters},
  volume  = {11},
  number  = {3},
  pages   = {3366--3373},
  year    = {2026}
}

@inproceedings{curtis2024tampura,
  author    = {Aidan Curtis and George Matheos and Nishad Gothoskar and Vikash Mansinghka and Joshua B. Tenenbaum and Tom{\'a}s Lozano-P{\'e}rez and Leslie Pack Kaelbling},
  title     = {Partially Observable Task and Motion Planning with Uncertainty and Risk Awareness},
  booktitle = {Robotics: Science and Systems (RSS)},
  year      = {2024}
}

@article{ma2025partiallyobservable,
  author  = {Yuhong Ma and Yeqing Yuan and Shaoquan Wu and Han Yuan},
  title   = {A Task and Motion Planning Framework for Partially Observable Household Manipulation Scenes},
  journal = {Advanced Intelligent Systems},
  year    = {2025},
  doi     = {10.1002/aisy.202400897}
}

@article{kim2026cocotamp,
  author  = {Yoonwoo Kim and Raghav Arora and Roberto Martin-Martin and Peter Stone and Ben Abbatematteo and Yoonchang Sung},
  title   = {Large-Language-Model-Guided State Estimation for Partially Observable Task and Motion Planning},
  journal = {arXiv preprint arXiv:2603.03704},
  year    = {2026}
}

@article{oloo2026vaptamp,
  author  = {Austine Oloo and Zainab Altaweel and Yohei Hayamizu and Peiqi Liu and Yan Ding and Saeid Amiri and Hao Yang and Andy Kaminski and Chad Esselink and Chris Paxton and Xiaohan Zhang and Shiqi Zhang},
  title   = {Robot Planning and Situation Handling with Active Perception},
  journal = {arXiv preprint arXiv:2604.26988},
  year    = {2026}
}

@inproceedings{curtis2022affordances,
  author    = {Aidan Curtis and Xiaolin Fang and Leslie Pack Kaelbling and Tom{\'a}s Lozano-P{\'e}rez and Caelan Reed Garrett},
  title     = {Long-Horizon Manipulation of Unknown Objects via Task and Motion Planning with Estimated Affordances},
  booktitle = {2022 IEEE International Conference on Robotics and Automation (ICRA)},
  year      = {2022},
  doi       = {10.1109/ICRA46639.2022.9812057}
}

@article{ding2022groundobjects,
  author  = {Yan Ding and Xiaohan Zhang and Xingyue Zhan and Shiqi Zhang},
  title   = {Learning to Ground Objects for Robot Task and Motion Planning},
  journal = {IEEE Robotics and Automation Letters},
  year    = {2022},
  doi     = {10.1109/LRA.2022.3155375}
}

@inproceedings{yuan2025robopoint,
  author    = {Wentao Yuan and Jiafei Duan and Valts Blukis and Wilbert Pumacay and Ranjay Krishna and Adithyavairavan Murali and Arsalan Mousavian and Dieter Fox},
  title     = {{RoboPoint}: A Vision-Language Model for Spatial Affordance Prediction in Robotics},
  booktitle = {Proceedings of the 8th Conference on Robot Learning},
  series    = {Proceedings of Machine Learning Research},
  volume    = {270},
  publisher = {PMLR},
  year      = {2025}
}

@misc{laurer2023building,
  title        = {Building Efficient Universal Classifiers with Natural Language Inference},
  author       = {Moritz Laurer and Wouter van Atteveldt and Andreu Casas and Kasper Welbers},
  year         = {2023},
  month        = dec,
  eprint       = {2312.17543},
  archivePrefix= {arXiv},
  primaryClass = {cs.CL},
  doi          = {10.48550/arXiv.2312.17543}
}

@inproceedings{huang2022zeroshot,
  title={Language Models as Zero-Shot Planners: Extracting Actionable Knowledge for Embodied Agents},
  author={Huang, Wenlong and Abbeel, Pieter and Pathak, Deepak and Mordatch, Igor},
  booktitle={Proceedings of the 39th International Conference on Machine Learning},
  pages={9118--9147},
  year={2022},
  volume={162},
  series={Proceedings of Machine Learning Research},
  publisher={PMLR}
}

@inproceedings{cheng2024yoloworld,
  title     = {YOLO-World: Real-Time Open-Vocabulary Object Detection},
  author    = {Cheng, Tianheng and Song, Lin and Ge, Yixiao and Liu, Wenyu and Wang, Xinggang and Shan, Ying},
  booktitle = {Proceedings of the IEEE/CVF Conference on Computer Vision and Pattern Recognition (CVPR)},
  pages     = {16901--16911},
  year      = {2024}
}

@article{ojha2026planwayeventtriggeredfoundationmodel,
  title={Plan Along the Way: Event-Triggered Foundation-Model Planning for TAMP Execution in Partially Observable Manipulation},
  author={Ojha, Puru and Vijayakumar, Narendhiran and Singhal, Nav and Varma, Girish and Thomas, Antony},
  journal={arXiv preprint arXiv:2608.28075},
  year={2026}
}

@article{kaelbling2013IJRR,
  title={Integrated task and motion planning in belief space},
  author={Kaelbling, Leslie Pack and Lozano-P{\'e}rez, Tom{\'a}s},
  journal={The International Journal of Robotics Research},
  volume={32},
  number={9-10},
  pages={1194--1227},
  year={2013},
  publisher={Sage Publications Sage UK: London, England}
}
\end{document}